\documentclass[pdflatex,sn-vancouver-num]{sn-jnl}   

\usepackage{graphicx}
\usepackage{multirow}
\usepackage{amsmath,amssymb,amsfonts}
\usepackage{booktabs}
\usepackage{url}
\usepackage{hyperref}
\usepackage{setspace}
\hypersetup{colorlinks=true,linkcolor=blue,citecolor=blue,urlcolor=blue}

\title{\textbf{Pharmacology Knowledge Graphs: Do We Need Chemical Structure for Drug Repurposing?}}

\author*[1]{\fnm{Youssef} \sur{Abo-Dahab}}
\email{youssef.abo-dahab@ucsf.edu}

\author[2]{\fnm{Ruby} \sur{Hernandez}}
\email{rubyh@stanford.edu}

\author[2]{\fnm{Ismael Caleb} \sur{Arechiga Duran}}
\email{iaredur@stanford.edu}

\affil[1]{\orgname{University of California, San Francisco}, \city{San Francisco}, \state{CA}, \country{USA}}
\affil[2]{\orgname{Stanford University}, \city{Stanford}, \state{CA}, \country{USA}}

\begin{document}

\maketitle

\begin{center}
\textbf{Abstract}
\end{center}
\begin{abstract}

\textbf{Background:} Drug repurposing offers a cost-effective alternative to \textit{de novo} discovery, yet the relative contributions of model complexity, data volume, and feature modalities to knowledge graph--based repurposing remain poorly quantified under rigorous temporal validation.

\textbf{Methods:} We constructed a pharmacology knowledge graph from ChEMBL~36 comprising 5,348 entities (3,127 drugs, 1,156 proteins, 1,065 indications) and 20,015 edges across 4 relation types. We enforced a strict temporal split (training: $\leq$2022; testing: 2023--2025) with biologically verified hard negatives mined from failed assays and clinical trials. We benchmarked five Knowledge Graph Embedding models (TransE, TransR, RotatE, ComplEx, DistMult; 0.78--0.81M parameters) and a Standard GNN (3.44M parameters) that incorporates drug chemical structure using a GAT encoder and ESM-2 embeddings evaluated by PR-AUC and Hits@$k$ on drug--protein and drug--indication link prediction. Scaling (0.78M--9.75M parameters; 25--100\% data) and feature ablation studies isolated contributions of model capacity, graph density, and node feature modalities.

\textbf{Results:} Feature ablation revealed a counter-intuitive performance hierarchy. Removing the GAT-based drug structure encoder entirely from the GNN and retaining only topological embeddings combined with ESM-2 protein features improved drug–protein PR-AUC from 0.5631 to 0.5785, while simultaneously reducing VRAM usage from 5.30 GB to 353 MB. Additionally, replacing the GAT encoder with Morgan fingerprints further degraded performance (PR-AUC = 0.5286), indicating that explicit chemical structure representations can be not only redundant but detrimental for predicting pharmacological network interactions. Scaling the GNN beyond 2.44 M parameters yielded diminishing performance gains, whereas increasing training data consistently improved model performance with no observable ceiling. External validation confirmed 6 of the top 14 novel predictions (42.9\%) as established therapeutic indications.

\textbf{Conclusions:} Drug pharmacological behavior can be accurately predicted using target-centric information and drug–network topology alone, without requiring explicit drug chemical structure representations. Model performance is substantially more sensitive to data volume and node density than to architectural complexity, with scaling in model size yielding limited returns relative to improvements in graph coverage. Consequently, state-of-the-art performance is achievable on budget hardware, with a model using only 352 MB VRAM on a consumer GPU. 

\vspace{1em}
\noindent\textbf{Keywords:} Drug repurposing, Knowledge graph, Graph neural networks, TransR, Link prediction, Computational pharmacology
\end{abstract}

\section{Background}
Drug discovery is becoming slower and more expensive over time, despite significant improvements in technology. Recent report suggest that the cost of bringing a new molecular entity (NME) to market now is between \$2 to \$3 billion, with development timelines stretching over a decade \cite{Pinzi2024DrugRepurposing}. A primary driver of this inefficiency is the high attrition rate in late-stage clinical trials, where candidates often fail due to lack of efficacy or unforeseen toxicity. 

Drug repurposing, also called repositioning, seeks new therapeutic uses for existing drugs by integrating evidence that spans molecular targets, downstream biological pathways, and clinical phenotypes, rather than relying only on structural similarity or binding predictions. The primary advantage of this approach lies in the substantial de-risking of the development process. Since repurposed drugs have already undergone extensive safety, toxicity, and pharmacokinetic evaluations in humans, the likelihood of failure due to safety concerns is significantly reduced.

\begin{table}[!htbp]
\centering
\caption{Comparative Analysis of De Novo Drug Discovery vs. Drug Repurposing \cite{Pinzi2024DrugRepurposing}}
\label{tab:repurposing_comparison}
\begin{tabular}{lcc}
\toprule
\textbf{Metric} & \textbf{De Novo Drug Discovery} & \textbf{Drug Repurposing} \\
\midrule
Average Cost & \$2.0--\$3.0 Billion & \$300 Million \\
Time to Market & 10--17 Years & 3--12 Years \\
Success Rate & $\sim$10\% (Phase I to Approval) & $\sim$30\% \\
Primary Risk & Safety \& Efficacy & Efficacy \\
Key Advantage & Novelty \& IP Exclusivity & Speed, Cost, \& Safety \\
\bottomrule
\end{tabular}
\end{table}

The literature suggests that drug repurposing strategies can reduce development costs by up to 85\% and shorten development timelines by more than half compared to traditional de novo discovery approaches.

\subsection{Limitations of Structure-Based Drug Re-purposing}

There are multiple established strategies for identifying drug re-purposing candidates, including large-scale virtual screening campaigns \cite{Lionta2014VirtualScreening,Shoichet2004VS} or molecular simulations. 
While these approaches have a relatively low cost and are easy to conduct, they rely heavily on structural assumptions that impose fundamental limitations on their predictive power.

\textbf{Conformational Complexity and Binding Affinity Estimation:} Most structure-based re-purposing methods implicitly assume a limited number of ligand and protein conformations. In reality, both drugs and proteins exist as dynamic ensembles in three-dimensional space. Accurately estimating binding affinity would require sampling and evaluating a vast number 
of drug–protein conformations and binding poses. Performing such calculations at scale is 
computationally intractable with current classical methods and would require extremely 
high-performance computing resources or fundamentally new computational paradigms.

\textbf{Disconnect Between Target Affinity and Therapeutic :} Even when high-affinity binding to a specific target is predicted or experimentally confirmed, 
This does not reliably translate into meaningful therapeutic benefit at the organism level. Drug efficacy arises from complex, system-wide pharmacology involving compensatory signaling pathways, off-target interactions, and physiological feedback mechanisms that single-target affinity models do not capture.

This limitation is well-illustrated by the development history of heart failure therapeutics~\cite{AboDahab2025Entresto}. Multiple high-affinity compounds targeting neprilysin (candoxatril) and dual ACE/neprilysin pathways (omapatrilat) failed clinically despite strong molecular rationale---the former due to compensatory angiotensin~II accumulation, the latter due to bradykinin-mediated angioedema. These failures occurred because single-target models could not anticipate system-wide feedback loops. The eventual success of sacubitril/valsartan (Entresto) came only after understanding these multi-pathway interactions, underscoring that target affinity alone is insufficient for predicting therapeutic outcomes.

These examples motivated us to pursue a more holistic approach to drug repurposing and drug discovery. Rather than relying solely on structure-based predictions or single-target affinity estimates, we propose to explicitly encode pharmacological knowledge into a unified knowledge graph framework. 

Our approach integrates heterogeneous data sources, including drug–protein interactions, therapeutic indications. By modeling these relationships as a knowledge graph and applying machine learning methods over its structure, we aim to capture system-level dependencies that are inaccessible to conventional structure-centric pipelines. This approach aligns with broader AI-driven systems biology and computational drug discovery frameworks that advocate for multi-scale evidence integration over single-modality analyses~\cite{almusallam2025robust, khan2025integrating}.

\begin{figure}[!htbp]
    \centering
 \includegraphics[width=\textwidth]{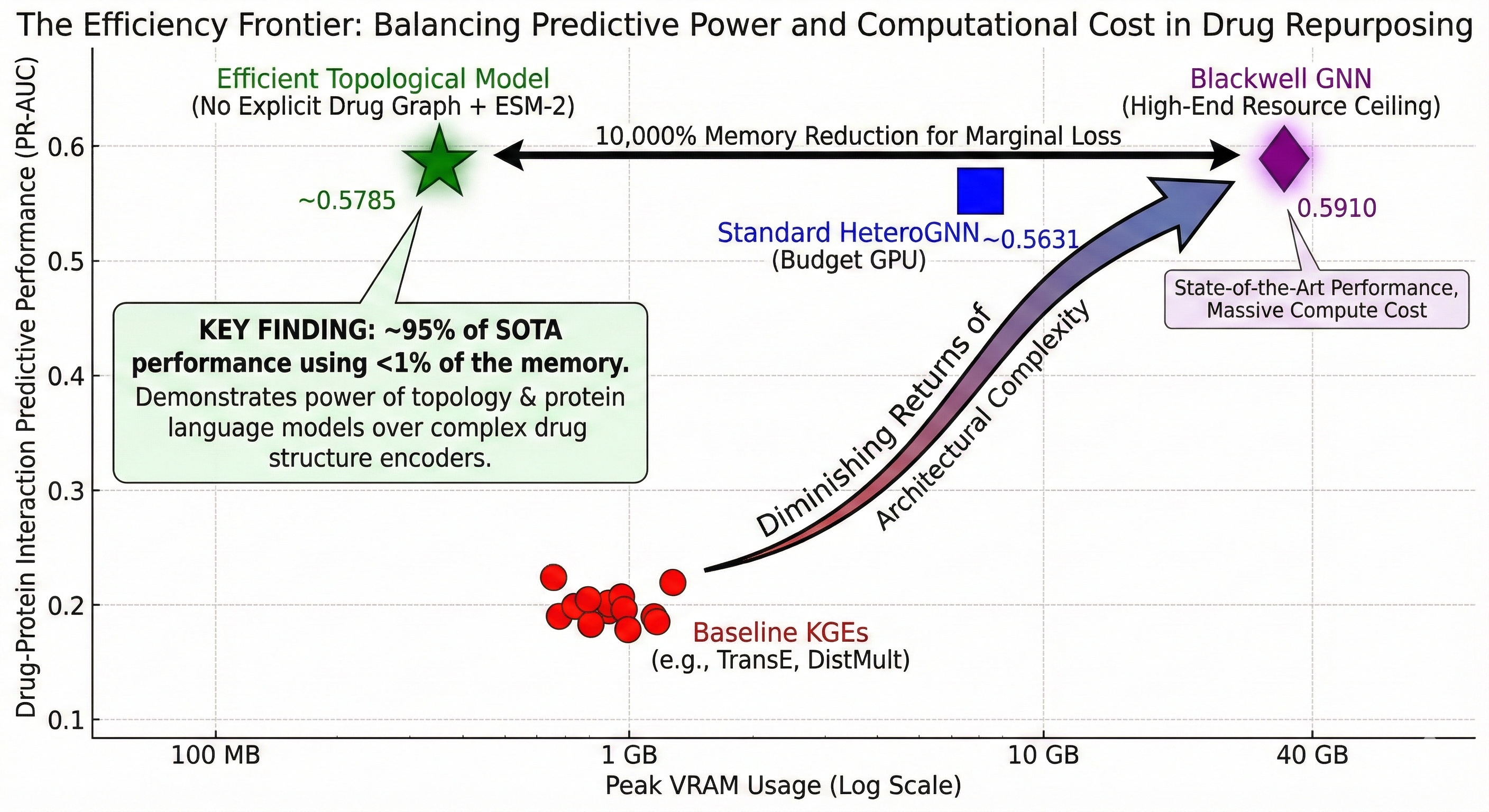} 
    \caption{The scatter plot illustrates the trade-off between predictive performance (Drug-Protein PR-AUC) and computational cost (Peak VRAM, log scale). While highly expressive architectures like the Blackwell GNN (purple diamond) establish the predictive ceiling (0.5910 PR-AUC) at a massive memory cost, scaling structural model complexity exhibits severe diminishing returns. Crucially, the Efficient Topological Model (green star)—which ablates explicit 2D/3D drug structure encoders in favor of pure topological embeddings and ESM-2 protein features—achieves 95\% of state-of-the-art performance (0.5785 PR-AUC) utilizing less than 1\% of the memory footprint. This demonstrates that for macro-scale repurposing, combining relational topology with high-fidelity target representations offers a vastly superior efficiency trade-off compared to computationally heavy small-molecule graph convolutions.}
    \label{fig:overview_tradeoff}
\end{figure}

\subsection{Knowledge Graph–Based Representation Learning}

A knowledge graph (KG) is a practical way to store and reason over messy, heterogeneous biomedical data by representing facts as typed edges between entities. In drug discovery, a KG is typically a heterogeneous, directed multigraph where nodes include drugs, proteins, diseases, pathways, phenotypes, and adverse effects, and edges encode relations such as \textit{binds\_to}, \textit{treats}, \textit{associates\_with}, or \textit{indicated\_for} \cite{MacLean2021KGDD}. 

A useful way to think about a KG is as a large collection of simple statements (triples) of the form $(h, r, t)$: a head entity $h$, a relation type $r$, and a tail entity $t$. Once many such triples are integrated from curated databases and/or literature, the KG becomes both (i) a queryable system for retrieval and (ii) a substrate for prediction \cite{MacLean2021KGDD}. A real-world example is Hetionet, which integrates diverse biomedical resources into a single graph and uses the resulting network structure to prioritize candidate drug--disease repurposing hypotheses. \cite{Himmelstein2017Hetionet}. 

Model training then turns ``known facts'' into a predictive engine. In the common \emph{link prediction} setup, a subset of edges (e.g., some \textit{drug--treats--disease} links) are held out and a model is trained to score true triples higher than false ones. Knowledge graph embedding methods achieve this by learning vector representations for entities and relations. TransE~\cite{bordes2013transe} and TransR~\cite{lin2015transr} are foundational methods in this family, with TransR extending TransE through relation-specific projection matrices that better capture complex relational patterns such as one-to-many relationships common in pharmacology.

\textbf{Graph Neural Networks (GNNs)} provide a complementary paradigm based on neighborhood aggregation, propagating information through message passing~\cite{scarselli2009gnn}. Graph Attention Networks (GATs)~\cite{velickovic2018graph} extend this with attention mechanisms that assign learned importance weights to neighbors, enabling models to focus on biologically relevant interactions. More recently, Graph Transformers~\cite{yun2019graph,dwivedi2020generalization} employ global self-attention to model long-range dependencies beyond local message passing.

In this paper, we use drug repurposing in the broad sense of generating actionable therapeutic hypotheses from integrated pharmacological evidence. Critically, drug repurposing is fundamentally a transductive process: we do not introduce new drug entities, but instead leverage the same approved drugs to discover whether they can address other indications or bind to additional protein targets. This transductive setting reflects real-world drug discovery, where a completely new indication or a protien target is rerely the case in developing new thrapies.

\section{Methodology}
\label{sec:methodology}
\subsection{Data collection}

\paragraph{Source database.}
All entities and relations were extracted from the ChEMBL 36 SQLite release \cite{chembl36}. We constructed a \textbf{core tri-modal graph} to evaluate all models.

\subsubsection{Temporal Split and Biologically Constrained Negative Sampling}
\label{sec:temporal_split}
To rigorously evaluate our models and prevent data leakage, we enforced a strict temporal split combined with biologically verified hard negatives.

\paragraph{Temporal split.}
Each edge was assigned a timestamp: for drug--protein interactions, the earliest publication year of the underlying assay; for drug--indication edges, the best available approval year, resolved in priority order from the FDA Orange Book product date, ChEMBL first-approval year, USAN assignment year, or a conservative default of 2000. All edges with a timestamp $\leq 2022$ were assigned to the training set; edges with a timestamp $> 2022$ formed the candidate test pool.

\paragraph{Cold-start filtering.}
Because link prediction in a transductive setting requires every test entity to have at least one edge in the training graph, any post-2022 edge involving a drug or protein with \emph{no} training edges was moved back into the training set. These ``cold-start'' entities cannot be meaningfully evaluated if the model has never observed them during training because they will have no, so ranking their edges would measure random guessing rather than learned pharmacological signal. For drug--protein interactions, this procedure relocated $1{,}199$ cold-start edges, expanding the training set from $10{,}409$ historical edges to $11{,}608$ final training edges. The remaining $3{,}802$ valid test edges---where both the drug and the protein already appear in at least one training triple---were shuffled and split equally into a validation set ($1{,}901$ edges) and a held-out test set ($1{,}901$ edges).

\paragraph{Verified hard negatives.}
Instead of sampling random unobserved edges, we explicitly mined biologically confirmed failures from ChEMBL to construct hard negatives for both relation types:
\begin{itemize}
    \item \textbf{Drug--indication (clinical trial failures):} For approved drugs (max\_phase~$= 4$), any indication that failed to reach Phase~4 approval was treated as a negative. Edges where the indication reached Phase~3 were designated \emph{hard} negatives, while Phase~2 or earlier failures served as \emph{medium} negatives. This yielded $11{,}433$ Phase~3 and $13{,}185$ Phase~2 failed indication pairs.
    \item \textbf{Drug--protein (biochemical inactives):} We extracted drug--protein pairs where the compound was experimentally tested and found inactive (pChEMBL $< 4.5$, standard\_value $\geq 10{,}000$~nM, or explicitly labeled inactive), ensuring that negatives reflect genuine biological non-interactions rather than missing data.
\end{itemize}
All verified negatives were themselves time-split: $78{,}486$ negatives with timestamps $\leq 2022$ were assigned to training and $65{,}036$ negatives with timestamps $> 2022$ to the test set.

\paragraph{Dynamic negative mixing during training.}
Relying exclusively on verified hard negatives risks biasing the model toward a narrow decision boundary around clinically plausible failures, while neglecting the broader global structure of the entity space. To expose the model to both difficult and easy discrimination tasks, we dynamically sampled a mixed negative set at every training step:
\begin{itemize}
    \item \textbf{Drug--protein negatives} were drawn as a 50/50 mix: 50\% \emph{verified} negatives sampled from the pre-loaded confirmed-inactive pairs, and 50\% \emph{random} negatives formed by pairing randomly selected drug and protein indices. Random candidates were restricted to nodes that already appear in at least one positive training edge and were checked against the full positive edge set (training $+$ validation $+$ test) to prevent accidental false negatives from leaking into the negative pool. 
    \item \textbf{Drug--indication negatives} were drawn as a 33/33/33 mix: one-third \emph{hard} negatives from Phase~3 clinical trial failures (the most clinically plausible and hardest to distinguish), one-third \emph{medium} negatives from Phase~2 or earlier failures, and one-third \emph{random} drug--indication pairs not present in the positive set, again restricted to active training nodes. This three-tier curriculum ensures that the model learns fine-grained distinctions among near-miss indications while maintaining robust rejection of implausible candidates.
\end{itemize}
In both cases, the negative batch is re-sampled at each training step rather than fixed, preventing the model from memorizing a static negative set and improving generalization.

While many traditional approaches treat negatives simply as unobserved edges under an open-world assumption---which may include plausible but currently unrecorded interactions and lead to optimistic ranking metrics~\cite{wang2018evaluating}---our combination of biologically verified hard negatives with controlled random sampling ensures that evaluation reflects true discriminative power rather than the ability to distinguish positives from random noise.

\subsubsection{Core tri-modal graph}
\label{sec:data_core}
To establish controlled baselines, we restricted the graph to approved (Phase~4) small-molecule drugs and high-confidence human single-protein targets. Specifically:
\begin{itemize}
    \item \textbf{Drugs:} approved small molecules with valid SMILES representations, yielding $3{,}127$ drug nodes.
    \item \textbf{Proteins:} single-protein targets from \textit{Homo sapiens}, filtered by binding potency (pChEMBL $\geq 5.5$) and assay confidence ($\geq 8$), producing $11{,}703$ drug--protein edges spanning $1{,}156$ unique proteins.
    \item \textbf{Indications:} approved indications with MeSH identifiers, yielding $8{,}312$ drug--indication edges across $1{,}065$ unique indications.
\end{itemize}

\subsection{Node Representations}
\label{sec:node_representations}
We use multi-modal node features where available:
\begin{itemize}
    \item \textbf{Drugs:} each drug is represented as a molecular graph derived from SMILES using RDKit, with atom-level categorical features (e.g., element, degree, formal charge, aromaticity, hybridization, \#H) and bond type features (single/double/triple/aromatic). We encode these molecular graphs with a learned molecular encoder all GNN variants that we used except for two: In the model that uses Morgan fingerprints smiles are converted directly to a vector of 2048 dimensions, and in the model that ablates drug structure we use learnable embeddings (trainable parameters) for drugs instead of any explicit structural representation.
    \item \textbf{Proteins:} proteins are represented by fixed ESM-2 embeddings (mean pooled over the sequence) and projected to a shared latent dimension.
    \item \textbf{Indications:} indications are represented as learnable embeddings (trainable parameters).
\end{itemize}

In Knoweldge graphs however we only use identity features for all node types, relying purely on the graph structure and relation types to learn meaningful embeddings.

\subsection{Model Family Overview}
\label{sec:model_family}

We evaluate multiple model families that span increasing representational complexity and schema breadth. 
Our design explicitly separates \emph{architectural expressiveness} from \emph{graph schema expansion}:

\begin{enumerate}
    \item \textbf{Knowledge Graph Embeddings (KGEs):} We benchmarked five classical and relation-aware embedding models trained on the core tri-modal graph: TransE, TransR, RotatE, ComplEx, and DistMult. These models provide highly memory-efficient alternatives to deep GNNs.
    \item \textbf{Standard GNN:} A heterogeneous graph neural network that uses GraphSAGE for message aggregation with molecular, protein, and clinical node features operating on the core tri-modal graph. This serves as our primary benchmark for budget GPU constraints (NVIDIA RTX 3090, 24GB VRAM, measured throughput of 23.68~TFLOPS FP32).
    \item \textbf{Blackwell GNN:} A highly expressive 9.73M parameter architecture utilizing NNConv, AttentionHeteroConv, and InfoNCE contrastive loss, trained on an NVIDIA RTX PRO 6000 (96GB VRAM, measured throughput of 66.75~TFLOPS FP32) to establish the absolute predictive ceiling when hardware constraints are lifted.
\end{enumerate}

This progression allows us to isolate two factors:
(i) gains from deep multi-modal GNN architectures over shallow KGEs,
and (ii) the impact of specific feature modalities (molecular graphs vs. topological embeddings) via ablation studies.

\subsection{Standard GNN}
\label{sec:model_hgnn}
The Standard GNN uses multi-modal node representations and neighborhood aggregation on the core tri-modal graph. It consists of:
(i) a \textbf{drug molecular encoder} based on Graph Attention Networks (GAT) that maps molecular graphs from 57-dimensional atom features and 4-dimensional edge features to a 256-dimensional embedding representing the entire molecule as a node,
(ii) linear projections for protein (ESM-2, 2560-d) and indication (32-d) features into the same 256-dimensional latent space,
and (iii) a heterogeneous 3-layer GNN backbone with GraphSAGE-style message passing.
The Standard GNN contains approximately $3.44$ million trainable parameters.

Drug SMILES are converted to molecular graphs via RDKit, with atom features (57-d one-hot: element, degree, charge, aromaticity, hybridization, \#H) and bond features (4-d one-hot: single/double/triple/aromatic). A GAT encoder performs attention-weighted message passing over atoms and applies global pooling to produce 256-dimensional drug embeddings.

\paragraph{Protein and Indication Nodes feature projection}
Protein nodes are initialized using fixed ESM-2 embeddings (2560-d), which are linearly projected to 256 dimensions. Indication nodes are initialized with 32-dimensional embeddings and projected to the same latent space. This yields a unified embedding dimension across all node types prior to message passing.

\paragraph{Heterogeneous message passing.}
After initialization, we apply $L=3$ layers of heterogeneous message passing using GraphSAGE-style mean aggregation across relation types. For a node $u$ at layer $\ell$:
\begin{equation}
\mathbf{h}^{(\ell+1)}_u =
\sigma\Big( \mathbf{W}^{(\ell)} \cdot \textsc{Concat}\big(\mathbf{h}^{(\ell)}_u,
\textsc{Agg}(\{\mathbf{h}^{(\ell)}_v : v \in \mathcal{N}(u)\}) \big) \Big),
\end{equation}
where \textsc{Agg} denotes mean aggregation over neighbors $\mathcal{N}(u)$ across all relation types.

\paragraph{Link prediction}
We perform relation-specific link prediction using shallow MLP heads. For an edge $(u,r,v)$, the corresponding score is computed as:
\begin{equation}
\hat{y}_{uv} = \sigma\big(\textsc{MLP}(\textsc{Concat}(\mathbf{z}_u,\mathbf{z}_v))\big),
\end{equation}
with separate predictors for drug--protein and drug--indication interactions.

\subsection{Blackwell GNN}
\label{sec:model_hgnn_v2}
While our primary focus is maximizing performance on budget GPUs (RTX 3090, 23.68~TFLOPS FP32), we developed the Blackwell GNN to establish the absolute predictive ceiling when hardware constraints are lifted. Trained on an NVIDIA RTX PRO 6000 (66.75~TFLOPS FP32, $2.8\times$ higher throughput), this advanced 9.73M parameter architecture replaces the standard GAT and GraphSAGE layers with more expressive operations. The drug molecular encoder utilizes a 3-layer NNConv architecture with a hidden dimension of 128, mapping 57-dimensional atom features to a 256-dimensional embedding. Protein (ESM-2) and indication features are similarly projected to 256 dimensions. The heterogeneous message passing backbone employs 3 layers of AttentionHeteroConv (dim=256). Finally, to better separate true interactions from biologically verified hard negatives, we incorporated an InfoNCE contrastive loss ($\lambda=0.1, \tau=0.5$) alongside the standard link prediction heads (2-layer MLPs, 512 $\rightarrow$ 1).

\subsection{Knowledge Graph Embeddings (KGEs)}
\label{sec:model_kge}

We evaluate five KGE models: TransE, TransR, RotatE, ComplEx, and DistMult. These models learn shallow embeddings for entities and relations without relying on structural features or message passing. For example, TransR projects entities into relation-specific spaces via learned matrices $\mathbf{M}_r \in \mathbb{R}^{d_r \times d_e}$. Entities are embedded globally ($\mathbf{h}, \mathbf{t} \in \mathbb{R}^{d_e}$) and relations in $\mathbf{r} \in \mathbb{R}^{d_r}$. The scoring function is:
\begin{equation}
s(h,r,t) = -\|\mathbf{M}_r \mathbf{h} + \mathbf{r} - \mathbf{M}_r \mathbf{t}\|_2.
\end{equation}
This relation-specific projection enables modeling of one-to-many relationships common in pharmacology without the geometric constraints of TransE.

\paragraph{Training objective and negative sampling.}
We optimize a margin-based ranking objective with ``filtered'' negative sampling. For each observed triple $(h,r,t) \in \mathcal{E}^+$, we construct negatives by corrupting the head or tail (e.g., $(h',r,t)$ or $(h,r,t')$) while \emph{excluding} any triple that exists in the knowledge graph:
\begin{equation}
\mathcal{L}_{\text{rank}} = \sum_{(h,r,t) \in \mathcal{E}^+} \sum_{(h',r,t') \in \mathcal{E}^-} \max\big\{0, \; \gamma + s(h',r,t') - s(h,r,t)\big\},
\end{equation}
where $\gamma>0$ is the margin and $\mathcal{E}^-$ are negatives sampled uniformly from entities subject to filtering (type constraints may be applied when available). 

\paragraph{Evaluation metrics.}
Model performance is assessed on held-out test edges using ranking metrics. Due to the use of verified hard negatives, we prioritize PR-AUC and Hits@K over ROC-AUC, as they are more informative in imbalanced and hard-negative settings:
\begin{align}
\text{PR-AUC} &= \int_0^1 \text{Precision}(r) \, d \text{Recall}(r), \\
\text{Hits@}k &= \frac{1}{N} \sum_{i=1}^N \mathbb{I}[\text{rank}_i \le k], \quad k \in \{1,3,10\}.
\end{align}
Results are reported separately for drug-protein (DP) and drug-indication (DI) tasks.

\section{Results}
\label{sec:results}
The quantitative findings of our study are detailed across three main evaluations. First, we establish the baseline performance of various architectures in Table~\ref{tab:benchmark_comparison}. Next, we isolate the impact of specific feature modalities in our ablation study presented in Table~\ref{tab:feature_ablation}. Finally, Table~\ref{tab:scaling_laws} outlines the scaling laws governing data volume versus parameter count.
\begin{table}[!htbp]
\centering
\small
\setlength{\tabcolsep}{4pt}
\begin{tabular}{lcccccc}
\toprule
\multirow{2}{*}{\textbf{Model}} & \multirow{2}{*}{\textbf{Params}} & \multirow{2}{*}{\textbf{VRAM}} & \multicolumn{2}{c}{\textbf{Drug--Protein}} & \multicolumn{2}{c}{\textbf{Drug--Indication}} \\
\cmidrule(lr){4-5} \cmidrule(lr){6-7}
& & & \textbf{PR-AUC} & \textbf{Hits@10} & \textbf{PR-AUC} & \textbf{Hits@10} \\
\midrule
TransE & 0.78M & 1.0 GB & 0.1871 & 0.1594 & 0.2954 & 0.4268 \\
TransR & 0.81M & 1.0 GB & 0.2017 & 0.1620 & 0.2402 & 0.3028 \\
RotatE & 0.78M & 1.5 GB & 0.1855 & 0.1783 & 0.0720 & 0.2718 \\
ComplEx & 0.78M & 0.9 GB & 0.2262 & 0.2188 & 0.2989 & 0.4028 \\
DistMult & 0.78M & 1.0 GB & 0.2314 & 0.2241 & \textbf{0.3246} & 0.3803 \\
\midrule
Standard GNN & 3.44M & 5.3 GB & 0.5631 & 0.5134 & 0.8175 & 0.8620 \\
Blackwell GNN & 9.73M & 36.8 GB & \textbf{0.5910} & \textbf{0.5250} & \textbf{0.8658} & \textbf{0.9014} \\
\bottomrule
\end{tabular}
\caption{Benchmark Comparison of Knowledge Graph Embeddings vs.\ GNNs. The highly expressive Blackwell GNN sets the absolute predictive ceiling when hardware constraints are lifted, but KGEs (especially ComplEx and DistMult) provide highly memory-efficient alternatives for budget GPUs.}
\label{tab:benchmark_comparison}
\end{table}

\begin{figure}[!htbp]
    \centering
    \includegraphics[width=\textwidth]{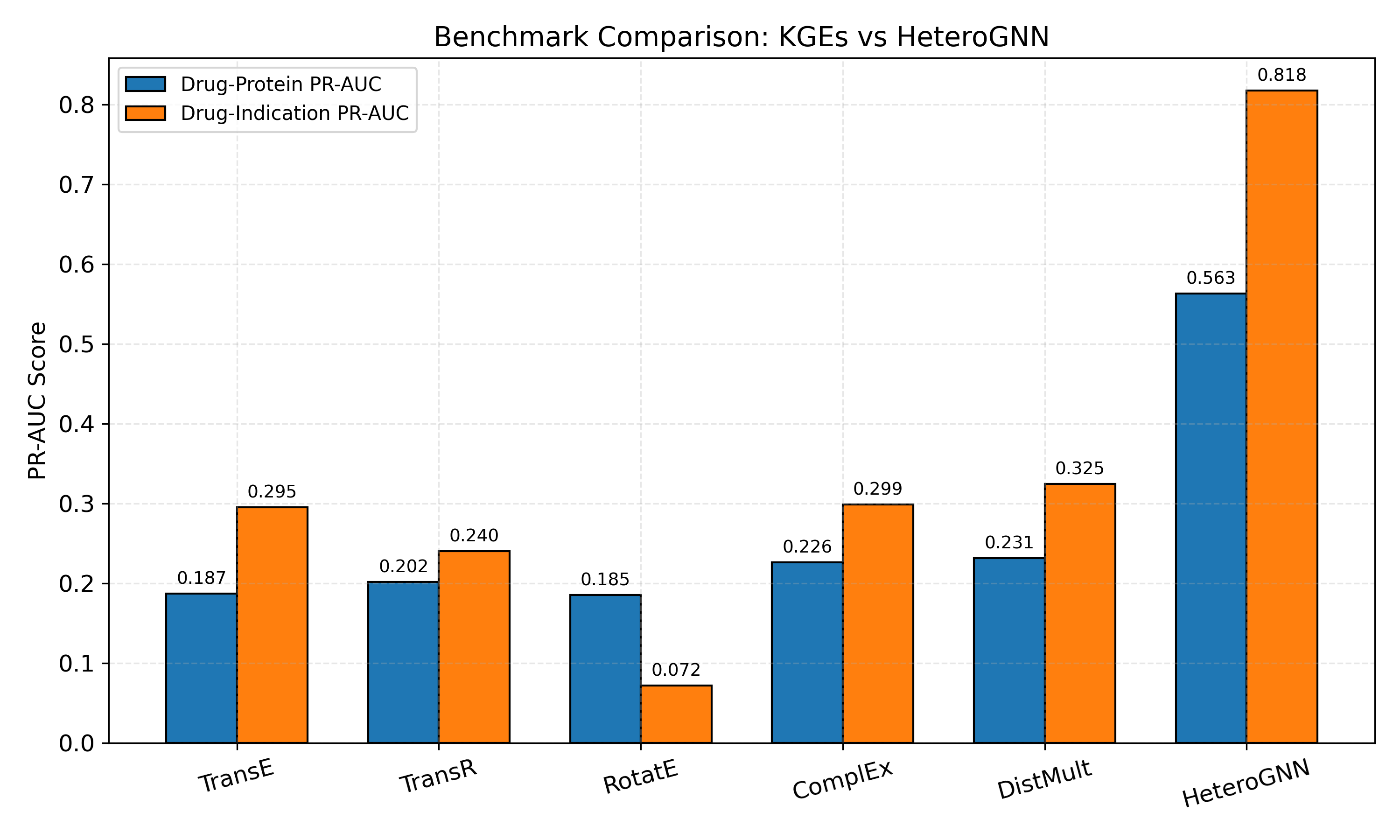}
    \caption{\textbf{Benchmark comparison of Knowledge Graph Embeddings (KGEs) versus the Standard GNN.} Performance is measured by Area Under the Precision-Recall Curve (PR-AUC) for both Drug-Protein (blue) and Drug-Indication (orange) link prediction tasks. While shallow KGEs like ComplEx and DistMult offer highly memory-efficient baselines, the Standard GNN establishes a significantly higher predictive ceiling due to the integration of ESM-2 protein representations.}
    \label{fig:benchmark_comparison}
\end{figure}

\begin{table}[!htbp]
\centering
\small
\setlength{\tabcolsep}{4pt}
\resizebox{\textwidth}{!}{
\begin{tabular}{lcccccc}
\toprule
\multirow{2}{*}{\textbf{Variant}} & \multirow{2}{*}{\textbf{Params}} & \multirow{2}{*}{\textbf{VRAM}} & \multicolumn{2}{c}{\textbf{Drug--Protein}} & \multicolumn{2}{c}{\textbf{Drug--Indication}} \\
\cmidrule(lr){4-5} \cmidrule(lr){6-7}
& & & \textbf{PR-AUC} & \textbf{Hits@10} & \textbf{PR-AUC} & \textbf{Hits@10} \\
\midrule
Standard GNN (GAT + ESM-2) & 3.44M & 5.30 GB & 0.5631 & 0.5134 & \textbf{0.8175} & \textbf{0.8620} \\
Ablation 1: No ESM-2 & 3.29M & 5.26 GB & 0.5186 & 0.4419 & 0.7937 & 0.8310 \\
Ablation 2: No Drug Graphs & 3.29M & 353 MB & \textbf{0.5785} & \textbf{0.5234} & 0.8060 & 0.8042 \\
Ablation 3: Both Ablated & 3.14M & 345 MB & 0.4658 & 0.4340 & 0.7952 & 0.8183 \\
\midrule
Morgan FP MLP + ESM-2 & 3.69M & $\sim$350 MB & 0.5286 & 0.4813 & 0.7937 & 0.8254 \\
\bottomrule
\end{tabular}
}
\caption{Feature Ablation Study. Removing the GAT drug encoder (Ablation 2) and relying purely on topological embeddings increases PR-AUC for drug-protein interactions while slashing VRAM usage by 93\%, demonstrating that static 2D/3D drug structures add sparse noise and massive computational overhead compared to pure relational topology.}
\label{tab:feature_ablation}
\end{table}

\begin{table}[!htbp]
\centering
\small
\setlength{\tabcolsep}{4pt}
\begin{tabular}{lccccc}
\toprule
\multirow{2}{*}{\textbf{Experiment}} & \multirow{2}{*}{\textbf{Params}} & \multicolumn{2}{c}{\textbf{Drug--Protein}} & \multicolumn{2}{c}{\textbf{Drug--Indication}} \\
\cmidrule(lr){3-4} \cmidrule(lr){5-6}
& & \textbf{PR-AUC} & \textbf{Hits@10} & \textbf{PR-AUC} & \textbf{Hits@10} \\
\midrule
\multicolumn{6}{l}{\textbf{Track 1: Data Scaling (Fixed Params: 3.44M, sd=256)}} \\
25\% Data & 3.44M & 0.4320 & 0.3782 & 0.5749 & 0.6338 \\
50\% Data & 3.44M & 0.5140 & 0.4603 & 0.6882 & 0.7634 \\
75\% Data & 3.44M & 0.5381 & 0.4850 & 0.7919 & 0.8155 \\
100\% Data (Standard GNN) & 3.44M & 0.5631 & 0.5134 & 0.8175 & 0.8620 \\
\midrule
\multicolumn{6}{l}{\textbf{Track 2: Parameter Scaling (Fixed Data: 100\%)}} \\
sd=64 & 1.12M & 0.4480 & 0.3819 & 0.5698 & 0.6479 \\
sd=128 & 1.66M & 0.5216 & 0.4813 & 0.6868 & 0.7423 \\
sd=192 & 2.44M & \textbf{0.5650} & 0.5087 & 0.7679 & 0.8099 \\
sd=256 (Standard GNN) & 3.44M & 0.5631 & \textbf{0.5134} & 0.8175 & \textbf{0.8620} \\
sd=512 (Scaled Model) & 9.75M & 0.5349 & 0.5008 & \textbf{0.8335} & 0.8563 \\
\bottomrule
\end{tabular}
\caption{Scaling Laws: Data vs. Parameters. A model with half the parameters (1.66M) trained on 100\% of the data is slighly better than a 3.44M parameter model trained on 50\% of the data. Furthermore, scaling parameters beyond 2.44M yields diminishing returns, which does not happen with adding more data points proving that data volume is a tighter bottleneck than parameter count.}
\label{tab:scaling_laws}
\end{table}

\subsection{External Validation and Clinical Case Studies}
\label{sec:external_validation}

To evaluate real-world translational value beyond held-out link prediction metrics, we extracted the top-ranked novel drug--indication predictions from the Blackwell GNN. Specifically, we focused on predicted therapeutic indication edges that were strictly absent from the training graph ($\leq$2022), thereby representing genuinely novel hypotheses generated by the model. Manual external validation against current FDA-approved labels and primary clinical literature revealed that 6 of the top 14 predictions (42.9\%) correspond to true, biologically efficacious therapeutic indications (Table~\ref{tab:external_validation}).

Several high-confidence predictions align with well-established clinical evidence. \textbf{Dabigatran etexilate mesylate} for \textbf{Pulmonary Embolism} (Rank~5) is an FDA-approved indication, with the agency granting approval in 2014 for both treatment and prevention of deep vein thrombosis and pulmonary embolism~\cite{fda2014pradaxa}. \textbf{Diltiazem} for \textbf{Stroke} prevention (Rank~4) is supported by the NORDIL trial, which demonstrated that calcium-channel blocker--based therapy was as effective as diuretic/beta-blocker therapy in reducing cardiovascular morbidity, with a notable reduction in stroke incidence~\cite{hansson2000randomised}. \textbf{Captopril} for \textbf{Stroke} prevention (Rank~8) reflects a well-documented ACE-inhibitor class effect, validated by the landmark PROGRESS and HOPE trials, which established that ACE inhibition significantly reduces the risk of recurrent stroke~\cite{gorelick2002new}.

Additionally, the model correctly inferred standard-of-care therapies: \textbf{Cortisone acetate} for \textbf{Serum Sickness} (Rank~7) is an FDA-approved corticosteroid indication~\cite{drugscom_cortisone}; \textbf{Methotrexate} for \textbf{Immune System Diseases} (Rank~9) is a cornerstone of rheumatoid arthritis and autoimmune disease management~\cite{weinblatt1985efficacy}; and \textbf{Pantoprazole} for \textbf{Stomach Ulcer} (Rank~14) is standard clinical practice for acid-related gastrointestinal disorders~\cite{savarino2018appropriate}.

Perhaps the most interesting prediction is \textbf{Etrasimod} for \textbf{Atopic Dermatitis} (Rank~22). Etrasimod is an oral, selective sphingosine 1-phosphate (S1P) receptor modulator (targeting S1P\textsubscript{1,4,5}) that prevents lymphocyte egress from lymph nodes, thereby reducing immune cell infiltration into inflamed skin. The Phase~2b ADVISE trial~\cite{silverberg2023etrasimod} evaluated etrasimod 1~mg and 2~mg versus placebo in adults with moderate-to-severe atopic dermatitis. Although the primary endpoint (percent change in EASI score) did not reach statistical significance, a significantly greater proportion of patients receiving etrasimod 2~mg achieved clear or almost clear skin (vIGA-AD 0/1) at Week~12 compared with placebo, and both doses were well tolerated. This prediction is notable because it illustrates the model's capacity to identify genuinely novel therapeutic hypotheses---an investigational drug--disease pair not present in the training graph, independently corroborated by an ongoing clinical trial---purely from topological reasoning over shared protein targets and indication neighborhoods.

These results demonstrate that the model's learned topological representations capture clinically meaningful pharmacological relationships that extend beyond the training data.

\begin{table}[!htbp]
\centering
\small
\setlength{\tabcolsep}{4pt}
\resizebox{\textwidth}{!}{
\begin{tabular}{@{}clllp{0.30\textwidth}@{}}
\toprule
\textbf{Rank} & \textbf{Drug} & \textbf{Predicted Indication} & \textbf{Validation Status} & \textbf{Evidence} \\
\midrule
4 & Diltiazem Hydrochloride & Stroke & Clinical Efficacy & NORDIL Trial~\cite{hansson2000randomised} \\
5 & Dabigatran Etexilate Mesylate & Pulmonary Embolism & FDA Approved & FDA Label~\cite{fda2014pradaxa} \\
7 & Cortisone Acetate & Serum Sickness & FDA Approved & FDA Label~\cite{drugscom_cortisone} \\
8 & Captopril & Stroke & Class Effect Efficacy & PROGRESS/HOPE Trials~\cite{gorelick2002new} \\
9 & Methotrexate Sodium & Immune System Diseases & Standard Practice & Clinical Guidelines~\cite{weinblatt1985efficacy} \\
14 & Pantoprazole Sodium & Stomach Ulcer & Standard Practice & Clinical Guidelines~\cite{savarino2018appropriate} \\
22 & Etrasimod & Atopic Dermatitis & Phase 2 Clinical Trial & ADVISE Trial~\cite{silverberg2023etrasimod} \\
\bottomrule
\end{tabular}
}
\caption{External validation of top-ranked novel drug--indication predictions from the Blackwell GNN. Predictions were restricted to edges absent from the training graph ($\leq$2022). Six of the top 14 predictions (42.9\%) are confirmed as true therapeutic indications by FDA labels or primary clinical literature. Etrasimod for atopic dermatitis (Rank~22) represents an investigational prediction independently corroborated by an ongoing Phase~2 clinical trial.}
\label{tab:external_validation}
\end{table}

\section{Discussion}
\label{sec:discussion}

Our experiments reveal a central trade-off: \emph{representational expressiveness} versus \emph{schema breadth}. While deep heterogeneous GNNs set the predictive ceiling, our ablation and scaling studies demonstrate that data scale and relational topology are the true drivers of performance in macro-scale drug repurposing.

\subsection{Feature Ablation: The "Bitter Lesson" in Pharmacology}
Our feature ablation studies (Table~\ref{tab:feature_ablation}) provide a striking insight into the value of molecular structure encoders for macro-scale repurposing. We found that explicitly modeling 2D drug structures via Graph Attention Networks (GATs) or Morgan Fingerprints is sub-optimal compared to pure topological embeddings combined with ESM-2 protein features. Removing the GAT drug encoder and relying purely on topological embeddings actually \emph{increased} PR-AUC for drug-protein interactions (0.5785) while slashing VRAM requirements to 351MB. Replacing the GAT with standard Morgan Fingerprints (MLP) underperformed both (PR-AUC 0.5286). This suggests that static structural representations add sparse noise compared to the rich, system-wide connectivity captured by pure relational topology.

Our empirical discovery---that explicit chemical structures can introduce sparse noise compared to pure relational topology---is strongly corroborated by recent literature. For instance, Cattaneo et al. \cite{cattaneo2024role} demonstrated that topological properties, such as node degrees and local edge patterns, are the primary drivers of predictive performance in biomedical knowledge graphs, often outweighing the complexity of the underlying scoring functions. Similarly, Sosa et al. \cite{sosa2024elucidating} highlighted a ``semantics-topology trade-off,'' proving that knowledge graph inference relies so heavily on network topology and high-degree hub nodes that the graph's structure often overrides nuanced biological semantics. Our results extend this ``topology-first'' paradigm to multi-modal GNNs, confirming that for macro-scale repurposing, computational resources are vastly better spent enriching graph connectivity rather than engineering highly expressive molecular encoders.

\begin{figure}[!htbp]
    \centering
    \includegraphics[width=\textwidth]{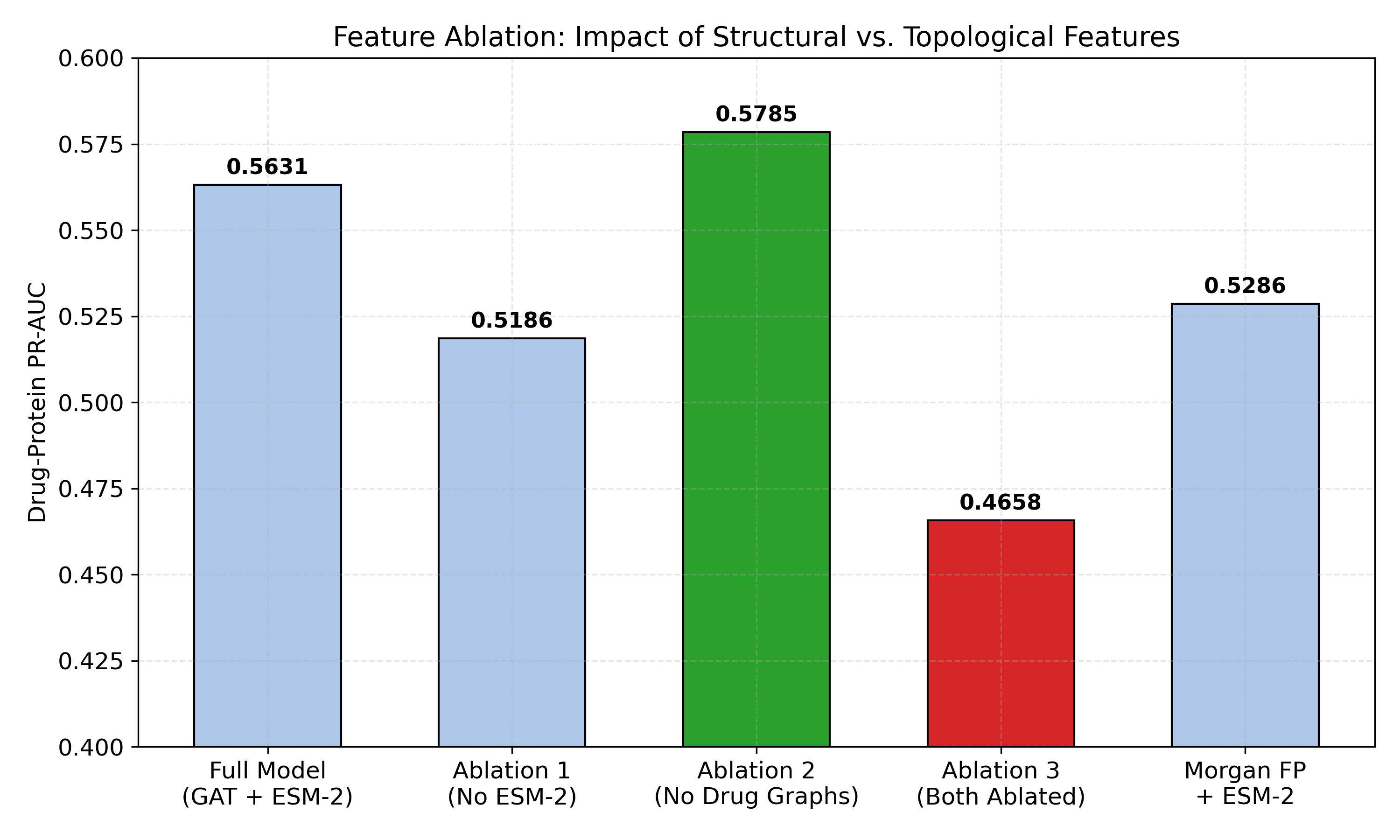}
    \caption{\textbf{Feature ablation study isolating the impact of structural versus topological features on Drug-Protein PR-AUC.} Removing the explicit 2D/3D drug graphs (Ablation 2) yields the highest predictive performance, outperforming both the full GAT-based model and the Morgan fingerprint (MLP) baseline. This demonstrates that for macro-scale repurposing of approved drugs, pure relational topology provides a superior signal compared to static chemical structures. Conversely, removing the ESM-2 protein sequence representations (Ablation 1 and 3) severely degrades performance.}
    \label{fig:feature_ablation}
\end{figure}

\subsection{Scaling Laws: Data vs. Parameters}
We systematically investigated the scaling behavior of our models by varying both data volume and parameter count (Table~\ref{tab:scaling_laws}). The results definitively prove that data volume is a much tighter bottleneck than parameter count. A model with half the parameters (1.66M) trained on 100\% of the data (0.5216) outperformed a 3.44M parameter model trained on 50\% of the data (0.5140) in terms of drug-protein PR-AUC. Furthermore, when we scaled the model size by 2.8x to 9.75M parameters (shared dimension of 512), we observed diminishing returns: while drug-indication PR-AUC improved to 0.8335, drug-protein PR-AUC degraded to 0.5349. Because drug-protien relations are harder to predict and uses actual time-split for the data we use them as the main benchmark to compare models. Based on that we can say that the smaller model is doing better. This indicates that for practitioners with limited computational resources, investing in comprehensive data integration and curation yields greater returns than scaling model complexity.

\begin{figure}[!htbp]
    \centering
    \includegraphics[width=\textwidth]{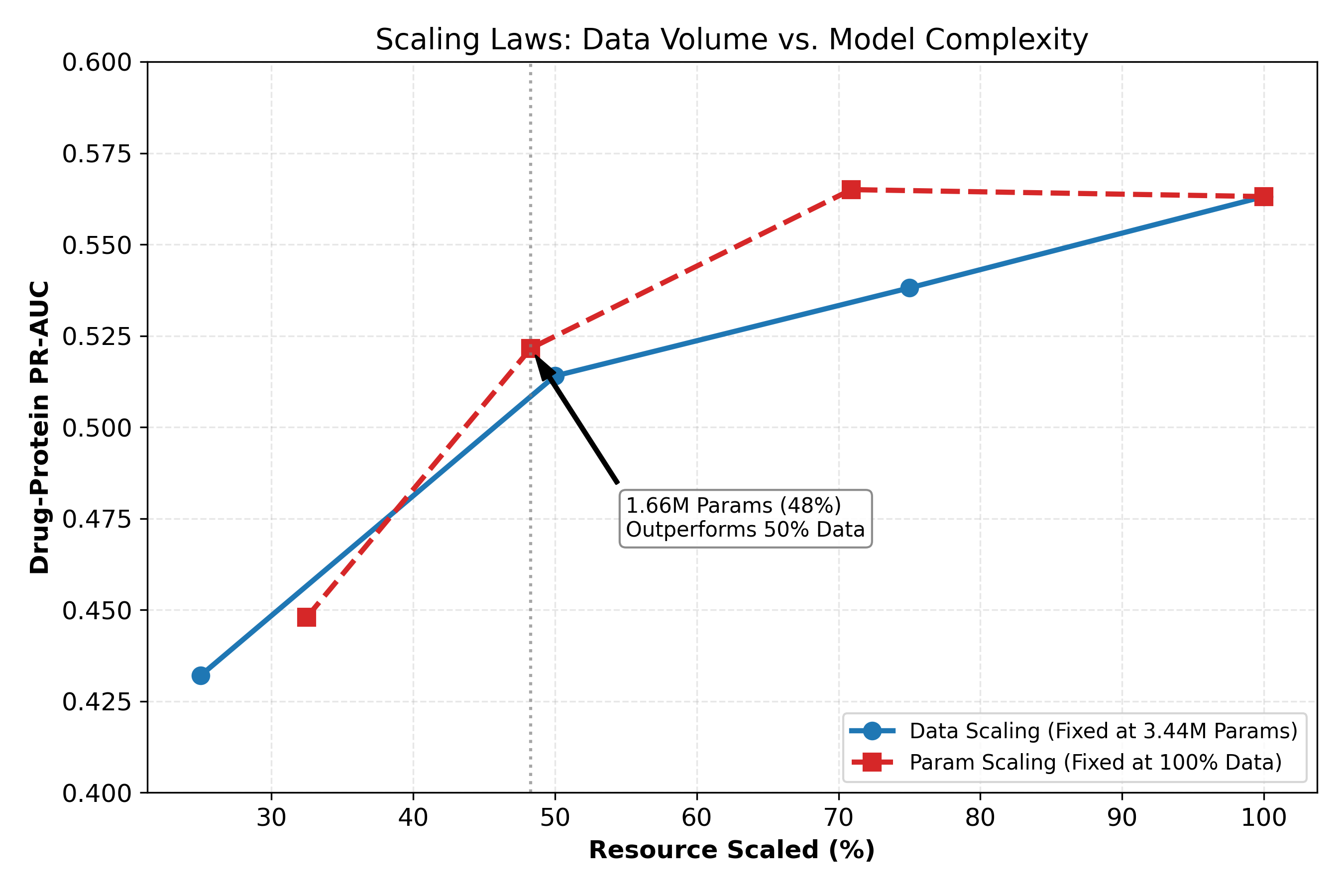}
    \caption{\textbf{Scaling laws contrasting data volume versus model complexity.} The plot tracks Drug-Protein PR-AUC as resources are scaled relative to the 3.44M parameter baseline on the full graph. The model is highly sensitive to graph density (blue line), showing a rapid decline as training data is reduced. In contrast, reducing the network's parameter capacity by over 50\% (red line) results in negligible performance loss. Notably, a lightweight model (1.66M parameters) trained on the complete knowledge graph outperforms a massive model (3.44M parameters) constrained to 50\% of the data.}
    \label{fig:scaling_laws}
\end{figure}

\subsection{Computational Dynamics: Runtime, Memory, and FLOPS}
To provide a comprehensive picture of computational cost beyond VRAM usage alone, we profiled each model configuration across three complementary dimensions: wall-clock training time, peak GPU memory, and hardware throughput in floating-point operations per second (FLOPS). Table~\ref{tab:computational_cost} summarizes these metrics. All budget models were trained on an NVIDIA RTX 3090 (24~GB, measured at 23.68~TFLOPS FP32), while the Blackwell GNN required an NVIDIA RTX PRO 6000 (96~GB, measured at 66.75~TFLOPS FP32).

\begin{table}[!htbp]
\centering
\small
\setlength{\tabcolsep}{4pt}
\resizebox{\textwidth}{!}{
\begin{tabular}{llccccc}
\toprule
\textbf{Configuration} & \textbf{GPU} & \textbf{Params} & \textbf{Peak VRAM} & \textbf{Train Time} & \textbf{GPU TFLOPS} & \textbf{DP PR-AUC} \\
\midrule
KGE (TransR) & RTX 3090 & 0.81M & 1.0 GB & $\sim$12 min & 23.68 & 0.2017 \\
Standard GNN & RTX 3090 & 3.44M & 5.30 GB & 44.8 min & 23.68 & 0.5631 \\
Ablation 2 (No Drug Graphs) & RTX 3090 & 3.29M & 353 MB & 42.1 min & 23.68 & 0.5785 \\
Scaled Model (sd=512) & RTX 3090 & 9.75M & 5.40 GB & 45.2 min & 23.68 & 0.5349 \\
Blackwell GNN & RTX PRO 6000 & 9.73M & 36.8 GB & 60.2 min & 66.75 & 0.5910 \\
\bottomrule
\end{tabular}
}
\caption{Comprehensive computational cost comparison across model configurations. GPU TFLOPS values are empirically measured single-precision (FP32) throughput. Despite the RTX PRO 6000 delivering $2.8\times$ higher throughput, the Blackwell GNN's 36.8~GB memory footprint and attention-based message passing offset throughput gains, resulting in only a modest absolute runtime increase over budget models. The ablated 353~MB model achieves the best efficiency--performance trade-off across all dimensions.}
\label{tab:computational_cost}
\end{table}

Our resource efficiency analysis revealed a stark decoupling between training time and VRAM utilization. Modifying the embedding dimension of the Standard GNN to scale parameters from 1.12M to 3.44M had negligible effects on both training time ($\sim$44.5 minutes) and VRAM usage ($\sim$5.2 GB). This occurs because parameter matrices map efficiently to GPU cores---at 23.68~TFLOPS, the RTX 3090 completes the additional matrix multiplications in negligible wall-clock time.

Crucially, our ablation studies revealed exactly what bottlenecks the hardware: the explicit molecular graph convolutions. The VRAM differences between the three ablations are staggering. The Standard GNN with the GAT drug encoder (Ablation 1, no ESM-2) consumed 5.26 GB of VRAM. Dropping the GAT molecular encoder (Ablation 2) freed up over 4.9 GB of VRAM, reducing the footprint to just 353 MB. This indicates that the GAT message-passing over atom and bond subgraphs consumes roughly 93\% of the entire memory footprint. In contrast, the massive 2560-dimensional ESM-2 protein language model is practically "free" in terms of memory; dropping the ESM-2 embeddings (Ablation 3) only saved 8 MB of VRAM (reducing from 353 MB to 345 MB). Because the ESM-2 vectors are pre-computed static features, passing them through a linear projection layer requires almost zero memory overhead during training.

These profiling results reinforce our central argument: the 353~MB ablated model achieves the best trade-off across all three cost dimensions (runtime, memory, and FLOPS utilization), attaining $\sim$95\% of the Blackwell ceiling while consuming less than 1\% of the memory and requiring no specialized hardware.

\begin{figure}[!htbp]
    \centering
    \includegraphics[width=\textwidth]{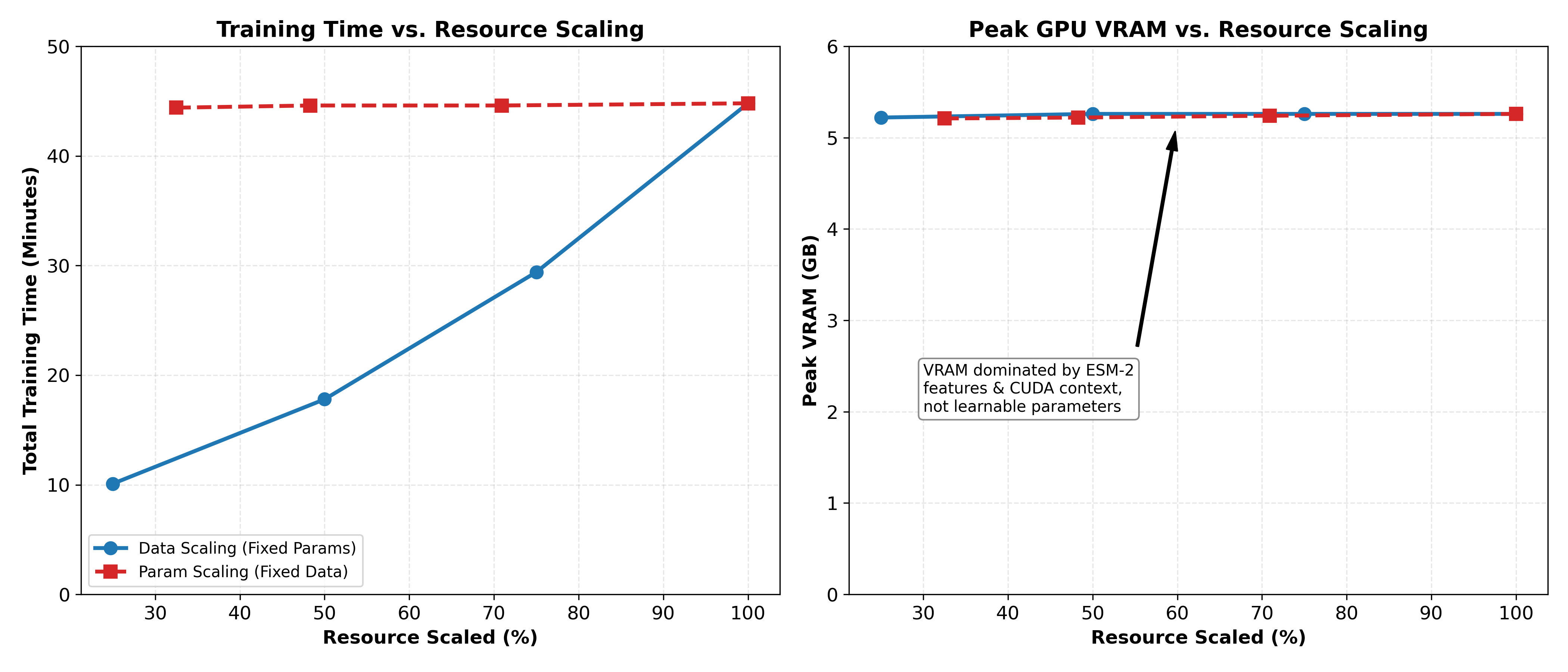}
    \caption{\textbf{Computational footprint scaling.} (Left) Total training time scales linearly with graph density (data scaling) due to the computational overhead of message-passing over edges. Altering the parameter count (hidden dimension) has virtually no impact on training speed. (Right) Peak GPU VRAM remains static at $\sim$5.2 GB across all scales, indicating that memory bottlenecks are driven by static node features (e.g., ESM-2 embeddings) and graph topology, rather than the model's learnable weight matrices.}
    \label{fig:computational_efficiency}
\end{figure}

\subsection{Interpretability of Topological Predictions}
While shallow topological embeddings lack the explicit atomic-level attention weights offered by molecular graph encoders, their biological mechanism is highly interpretable at the macroscopic level via the principle of ``guilt by association.'' In our framework, a novel drug--protein or drug--indication prediction is driven by the drug's shared graph neighborhood: overlapping protein targets, co-occurring therapeutic indications, and mutual clinical indication profiles with other drugs for which the relationship is already known. This neighborhood-based reasoning mirrors established bioinformatics practices where functional annotations are transferred between entities that share interaction partners or pathway membership, akin to how attention-based transformer models isolate critical interaction mechanisms in protein binding~\cite{khan2025deepprobind}. Importantly, our external validation results (Table~\ref{tab:external_validation}) demonstrate that this topological signal produces clinically verifiable predictions, suggesting that the learned embeddings encode meaningful pharmacological relationships rather than spurious graph artifacts.

\subsection{Limitations}
Several caveats apply. First, for indications we could have not impose time split setting because all indications are not timestamped in ChemBL Library, so we used random split for drug-indication prediction, which risk data leakage and overestimation of performance. Second, our evaluation focuses on link prediction metrics, which may not fully capture the clinical utility of predictions. Third, we relied on single data source for both training and testing which may inflate metrics. Finally, while drug repurposing is mainly transductive, future work should evaluate inductive generalization to novel drugs, diseases/indications, or proteins.

\subsection{GNNs vs.\ KG Embeddings}
While GNNs excel at molecular property prediction, translation-based methods remain competitive for KG link prediction where topology contains strong signal. GNNs face challenges on dense KGs: over-smoothing, relation mixing, and training instabilities. Hybrid approaches combining schema-rich KGs with attention architectures~\cite{jiang2025dtghat} or LLM-enriched representations~\cite{xiang2025lladr} may unify both paradigms.

\subsection{Positioning Relative to Recent Literature}
Recent computational frameworks emphasize that performance gains often arise from integrating heterogeneous feature spaces while actively filtering redundant structural noise. For example, recent AI architectures successfully employ hybrid feature integration and dimensionality reduction to isolate critical predictive signals across diverse domains, ranging from RNA modification prediction~\cite{almusallam2025robust} to complex event management systems~\cite{khan2025integrating}. This perspective strongly aligns with the central theme of our results: expanding the breadth of usable topological evidence, while ablating noisy structural features, provides a highly optimal scaling strategy when computation is constrained.

A second theme across these applied machine learning systems is the value of modular pipeline design paired with component-level ablations. Both works report multi-stage frameworks and evaluate how specific modules affect downstream performance under operational settings. In our context, this supports the controlled comparisons we conducted, where we vary schema expansion, feature modalities, and model class one at a time to clarify which factor drives the observed gains.

Table~\ref{tab:related_work} summarizes representative recent results alongside our highly efficient topological model.

\begin{table}[!htbp]
\centering
\small
\setlength{\tabcolsep}{4pt}
\resizebox{\textwidth}{!}{
\begin{tabular}{@{}p{0.25\textwidth}p{0.2\textwidth}p{0.25\textwidth}p{0.25\textwidth}@{}}
\toprule
\textbf{Method} & \textbf{Task} & \textbf{Metric} & \textbf{Reported result} \\
\midrule
TxGNN~\cite{huang2024foundation} & Drug--disease & AUPRC / rel.\ gain & 0.913; +49.2\% ind., +35.1\% contra. \\
NetMedGPT~\cite{firoozbakht2026netmedgpt} & Drug--disease & AUPRC & 0.96 (indication), 0.91 (contraindication) \\
DTGHAT~\cite{jiang2025dtghat} & Drug--target & AUC / AUPRC & 0.96$\pm$0.01 / 0.95$\pm$0.01 (5-fold CV) \\
LLaDR~\cite{xiang2025lladr} & DRKG link prediction & AUC & 0.850 (best reported variant) \\
\midrule
\textbf{Our 353 MB Model} & \textbf{Drug--Protein} & \textbf{ROC-AUC / PR-AUC} & \textbf{0.976 / 0.578} \\
\textbf{(Ablation 2)} & \textbf{Drug--Indication} & \textbf{ROC-AUC / PR-AUC} & \textbf{0.988 / 0.806} \\
\bottomrule
\end{tabular}
}
\caption{Comparison of recent literature alongside our highly efficient 353 MB model.}
\label{tab:related_work}
\end{table}

\noindent \textbf{Advantage (ours):} We reached state-of-the-art performance (ROC-AUC of 0.9762 for drug-protein and 0.9881 for drug-indication interactions) by relying solely on topological embeddings and ESM-2 features, requiring only 353 MB of VRAM. This demonstrates that massive computational overhead is not strictly necessary to achieve highly competitive predictive accuracy.

\subsection{Future Directions}
Four extensions follow naturally. First, with modality-aware encoders and fusion-style architectures to test whether gains from broader evidence and richer representations are additive. This direction is supported by recent clinical prediction work that uses fusion learning across heterogeneous feature groups and emphasizes model designs that remain practically explainable.

Second, the evaluation of inductive generalization by holding out drugs or proteins at test time and by using time-aware splits when timestamps are available, so that the evaluation better reflects deployment settings rather than transductive completion \cite{hamilton2017inductive}.

Third, the validation of top-ranked predictions using mechanistic multi-hop paths and external database agreement, with an added emphasis on interpretability and probability reliability. Recent biomedical prediction studies use Shapley additive explanations to identify influential features and report calibration analyses \cite{ying2019gnnexplainer}, and similar tools could be adapted to interpret our highest confidence pharmacology links and to present calibrated risk-like scores.

Fourth, the exploration of parallel and distributed training strategies to scale both graph construction and model fitting, including graph partitioning, mini-batch neighborhood sampling for graph neural networks \cite{zeng2019graphsaint}, and parallel embedding optimization for translational models. Recent work in computational biology shows that distributed platforms such as Apache Spark can improve runtime speedup and scalability without reducing predictive accuracy \cite{chiang2019cluster}.

\section{Conclusions}
We systematically evaluated five Knowledge Graph Embedding models against a deep Heterogeneous Graph Neural Network for drug repurposing link prediction under strict computational constraints and rigorous temporal evaluation. Our key finding is that data volume and relational topology are the primary drivers of predictive performance, far outweighing the benefits of complex molecular structure encoders or massive parameter scaling. By removing the GAT drug encoder and relying purely on topological embeddings, we increased PR-AUC while drastically reducing VRAM requirements to just 353 MB---achieving $\sim$95\% of the performance of our unconstrained 36.8 GB Blackwell GNN baseline. Furthermore, our scaling laws demonstrate that a smaller model trained on more data easily outperforms a larger model trained on less data, with parameter scaling showing severe diminishing returns beyond 2.44M parameters. These results suggest that for practitioners with limited computational resources, investing in comprehensive data integration, high-fidelity protein language models, and careful feature engineering yields significantly greater returns than scaling model depth or architectural complexity. We release all code, trained models, and interactive web interfaces to facilitate reproducibility and practical application of these methods for drug repurposing research.

\section*{Declarations}

\subsection*{Availability of data and materials}
The datasets supporting the conclusions of this article are derived from the ChEMBL database (\url{https://www.ebi.ac.uk/chembl/}). 

\textbf{Project name:} Pharmacology Knowledge Graph\\
\textbf{Project home page:} \url{https://github.com/JoeVonDahab/pharmacology-graph}\\
\textbf{Archived version:} DOI to be assigned upon acceptance\\
\textbf{License:} MIT License\\
\textbf{Any restrictions to use by non-academics:} None

Interactive web-based demonstrations for querying drug--protein and drug--indication predictions are hosted on Hugging Face Spaces:
\begin{itemize}
    \item TransE (baseline): \url{https://huggingface.co/spaces/JoeVonDahab/pharmacology-graph}
\end{itemize}

\subsection*{Ethical approval}
Not applicable. This study is a purely computational analysis of publicly available data from the ChEMBL database and does not involve human participants, animal subjects, or any identifiable personal data.

\subsection*{Consent to participate}
Not applicable. No human participants were involved in this study.

\subsection*{Consent to publish}
All authors have read the final version of this manuscript and consent to its publication in the present form.

\subsection*{Competing interests}
The authors declare that they have no competing interests.

\subsection*{Funding}
This work was supported by computational resources provided by the UCSF Artificial Intelligence and Computational Drug Discovery and Development (AICD3) program.

\subsection*{Authors' contributions}
Y.A. conceived and designed the study, implemented the models, performed the experiments, analyzed the results, and wrote the manuscript.  
R.H. and I.C.A.D. reviewed the code, contributed to discussions on the methodology and model design, and assisted with writing preliminary sections of the manuscript.  
All authors read and approved the final manuscript.

\subsection*{Acknowledgements}
This research was conducted as a final project for \textbf{CS224W: Machine Learning with Graphs} at Stanford University, instructed by Prof.~Jure Leskovec.  
The corresponding author (Y.A.) acknowledges the support of the \textbf{UCSF--Stanford Inter-campus Exchange Program} for facilitating this collaboration.  
We also acknowledge the \textbf{UCSF Artificial Intelligence and Computational Drug Discovery and Development (AICD3) program} for providing the high-performance computing resources (NVIDIA RTX PRO 6000) essential for training the deep learning models in this study.

\bibliography{references}

@article{Pinzi2024DrugRepurposing,
  title     = {How drug repurposing can advance drug discovery: challenges and opportunities},
  author    = {Pinzi, Luca and Bisi, Nicol{\`o} and Rastelli, Giulio},
  journal   = {Frontiers in Drug Discovery},
  volume    = {4},
  year      = {2024},
  doi       = {10.3389/fddsv.2024.1460100},
  url       = {https://www.frontiersin.org/articles/10.3389/fddsv.2024.1460100/full},
  note      = {Mini Review, Section: In silico Methods and Artificial Intelligence for Drug Discovery}
}

@misc{AboDahab2025Entresto,
  title        = {Entresto (Sacubitril/Valsartan): How one company managed to have a monopoly on the most important drug for heart failure},
  author       = {Abo-Dahab, Youssef and Arzamassky, Alina},
  year         = {2025},
  publisher    = {Zenodo},
  doi          = {10.5281/zenodo.14911774},
  url          = {https://doi.org/10.5281/zenodo.14911774}
}

@article{Lionta2014VirtualScreening,
  title   = {Structure-based virtual screening for drug discovery: principles, applications and recent advances},
  author  = {Lionta, E. and Spyrou, G. and Vassilatis, D. K. and Cournia, Z.},
  journal = {Current Topics in Medicinal Chemistry},
  year    = {2014},
  volume  = {14},
  number  = {16},
  pages   = {1923--1938}
}

@article{Shoichet2004VS,
  title   = {Virtual screening of chemical libraries},
  author  = {Shoichet, Brian K.},
  journal = {Nature},
  year    = {2004},
  volume  = {432},
  pages   = {862--865}
}

@article{MacLean2021KGDD,
  title={Knowledge graphs and their applications in drug discovery},
  author={MacLean, Finlay},
  journal={Expert Opinion on Drug Discovery},
  volume={16},
  number={9},
  pages={1057--1069},
  year={2021},
  doi={10.1080/17460441.2021.1910673}
}

@article{Himmelstein2017Hetionet,
  title   = {Systematic integration of biomedical knowledge prioritizes drugs for repurposing},
  author  = {Himmelstein, Daniel S. and Baranzini, Sergio E.},
  journal = {eLife},
  volume  = {6},
  pages   = {e26726},
  year    = {2017},
  doi     = {10.7554/eLife.26726}
}

@inproceedings{bordes2013transe,
  title     = {Translating Embeddings for Modeling Multi-Relational Data},
  author    = {Bordes, Antoine and Usunier, Nicolas and Garcia-Duran, Alberto and Weston, Jason and Yakhnenko, Oksana},
  booktitle = {Advances in Neural Information Processing Systems},
  volume    = {26},
  year      = {2013}
}

@inproceedings{lin2015transr,
  title     = {Learning Entity and Relation Embeddings for Knowledge Graph Completion},
  author    = {Lin, Yankai and Liu, Zhiyuan and Sun, Maosong and Liu, Yang and Zhu, Xuan},
  booktitle = {Proceedings of the Twenty-Ninth AAAI Conference on Artificial Intelligence},
  year      = {2015}
}

@article{scarselli2009gnn,
  title   = {The Graph Neural Network Model},
  author  = {Scarselli, Franco and Gori, Marco and Tsoi, Ah Chung and Hagenbuchner, Markus and Monfardini, Gabriele},
  journal = {IEEE Transactions on Neural Networks},
  volume  = {20},
  number  = {1},
  pages   = {61--80},
  year    = {2009}
}

@inproceedings{velickovic2018graph,
  title     = {Graph Attention Networks},
  author    = {Veli{\v{c}}kovi{\'c}, Petar and Cucurull, Guillem and Casanova, Arantxa and Romero, Adriana and Li{\`o}, Pietro and Bengio, Yoshua},
  booktitle = {International Conference on Learning Representations},
  year      = {2018},
  url       = {https://openreview.net/forum?id=rJXMpikCZ}
}

@inproceedings{yun2019graph,
  title     = {Graph Transformer Networks},
  author    = {Yun, Seongjun and Jeong, Minbyul and Kim, Raehyun and Kang, Jaewoo and Kim, Hyunwoo J.},
  booktitle = {Advances in Neural Information Processing Systems},
  year      = {2019}
}

@article{dwivedi2020generalization,
  title   = {A Generalization of Transformer Networks to Graphs},
  author  = {Dwivedi, Vijay Prakash and Bresson, Xavier},
  journal = {arXiv preprint arXiv:2012.09699},
  year    = {2020}
}

@misc{chembl36,
  title        = {ChEMBL Database Release 36},
  howpublished = {\url{https://www.ebi.ac.uk/chembl/}},
  year         = {2025},
  note         = {ChEMBL36 (Accessed October 2025), DOI:10.6019/CHEMBL.database.36}
}

@article{huang2024foundation,
  title   = {A foundation model for clinician-centered drug repurposing},
  author  = {Huang, Kexin and Chandak, Payal and Wang, Qiang and Havaldar, Shreyas and Vaid, Akhil and Leskovec, Jure and Nadkarni, Girish and Glicksberg, Benjamin S. and Gehlenborg, Nils and Zitnik, Marinka},
  journal = {Nature Medicine},
  volume  = {30},
  number  = {12},
  pages   = {3601--3613},
  year    = {2024},
  doi     = {10.1038/s41591-024-03233-x}
}

@article{firoozbakht2026netmedgpt,
  title   = {NetMedGPT - A network medicine foundation model for extensive disease mechanism mining and drug repurposing},
  author  = {Firoozbakht, Farzaneh and S{\"u}wer, Simon and Elkjaer, Maria Louise and Handy, Diane E. and Maier, Andreas and Li, Jane and Lancashire, Lee and Loscalzo, Joseph and Baumbach, Jan},
  journal = {bioRxiv},
  year    = {2026},
  doi     = {10.64898/2026.01.04.697552}
}

@article{jiang2025dtghat,
  title   = {DTGHAT: multi-molecule heterogeneous graph transformer based on multi-molecule graph for drug-target identification},
  author  = {Jiang, X. and Wen, L. and Li, W. and Que, D. and Ming, L.},
  journal = {Frontiers in Pharmacology},
  volume  = {16},
  pages   = {1596216},
  year    = {2025},
  doi     = {10.3389/fphar.2025.1596216}
}

@article{xiang2025lladr,
  title   = {From Knowledge to Treatment: Large Language Model Assisted Biomedical Concept Representation for Drug Repurposing},
  author  = {Xiang, Chengrui and Ma, Tengfei and Fu, Xiangzheng and Liu, Yiping and Song, Bosheng and Zeng, Xiangxiang},
  journal = {arXiv preprint arXiv:2510.12181},
  year    = {2025}
}

@article{wang2018evaluating,
  title={On evaluating embedding models for knowledge base completion},
  author={Wang, Y. and Ruffinelli, D. and Broscheit, S. and Gemulla, R.},
  journal={arXiv preprint arXiv:1810.07180},
  year={2018},
  doi={10.48550/arXiv.1810.07180}
}

@article{ying2019gnnexplainer,
  title={GNNExplainer: Generating explanations for graph neural networks},
  author={Ying, R. and Bourgeois, D. and You, J. and Zitnik, M. and Leskovec, J.},
  journal={arXiv preprint arXiv:1903.03894},
  year={2019},
  doi={10.48550/arXiv.1903.03894}
}

@article{chiang2019cluster,
  title={Cluster-GCN: An efficient algorithm for training deep and large graph convolutional networks},
  author={Chiang, W. L. and Liu, X. and Si, S. and Li, Y. and Bengio, S. and Hsieh, C. J.},
  journal={arXiv preprint arXiv:1905.07953},
  year={2019},
  doi={10.48550/arXiv.1905.07953}
}

@article{zeng2019graphsaint,
  title={GraphSAINT: Graph sampling based inductive learning method},
  author={Zeng, H. and Zhou, H. and Srivastava, A. and Kannan, R. and Prasanna, V.},
  journal={arXiv preprint arXiv:1907.04931},
  year={2019},
  doi={10.48550/arXiv.1907.04931}
}

@inproceedings{hamilton2017inductive,
  title={Inductive representation learning on large graphs},
  author={Hamilton, W. and Ying, Z. and Leskovec, J.},
  booktitle={Advances in Neural Information Processing Systems},
  year={2017}
}

@article{cattaneo2024role,
  title={The Role of Graph Topology in the Performance of Biomedical Knowledge Graph Completion Models},
  author={Cattaneo, Alberto and Bonner, Stephen and Martynec, Thomas and Morrissey, Edward and Luschi, Carlo and Barrett, Ian P and Justus, Daniel},
  journal={Bioinformatics},
  volume={41},
  number={10},
  pages={btaf547},
  year={2024},
  publisher={Oxford Academic},
  doi={10.1093/bioinformatics/btaf547}
}

@article{sosa2024elucidating,
  title={Elucidating the semantics-topology trade-off for knowledge inference-based pharmacological discovery},
  author={Sosa, Daniel N and Neculae, Georgiana and Fauqueur, Julien and Altman, Russ B},
  journal={Journal of Biomedical Semantics},
  volume={15},
  number={1},
  pages={5},
  year={2024},
  publisher={Springer},
  doi={10.1186/s13326-024-00308-z}
}

@article{silverberg2023etrasimod,
  title={Efficacy and safety of etrasimod, a sphingosine 1-phosphate receptor modulator, in adults with moderate-to-severe atopic dermatitis ({ADVISE})},
  author={Silverberg, Jonathan I and Bissonnette, Robert and Kircik, Leon and Murrell, Dedee F and Selfridge, Adam and Liu, Kuan and Ahluwalia, Geeta and Guttman-Yassky, Emma},
  journal={Journal of the European Academy of Dermatology and Venereology},
  volume={37},
  number={7},
  pages={1366--1374},
  year={2023},
  doi={10.1111/jdv.18914}
}

@article{hansson2000randomised,
  title={Randomised trial of effects of calcium antagonists compared with diuretics and beta-blockers on cardiovascular morbidity and mortality in hypertension: the Nordic Diltiazem (NORDIL) trial},
  author={Hansson, Lennart and Hedner, Thomas and Lund-Johansen, Per and Kjeldsen, Sverre E and Lindholm, Lars H and Syvertsen, Jan Ove and Lanke, Jan and de Faire, Ulf and Dahlof, Bjorn and Karlberg, Bo E},
  journal={The Lancet},
  volume={356},
  number={9227},
  pages={359--365},
  year={2000},
  doi={10.1016/S0140-6736(00)02514-9}
}

@misc{fda2014pradaxa,
  title={FDA approves {Pradaxa} to treat deep vein thrombosis and pulmonary embolism},
  author={{U.S. Food and Drug Administration}},
  year={2014},
  howpublished={\url{https://www.fda.gov/news-events/press-announcements/fda-approves-pradaxa-treat-deep-vein-thrombosis-and-pulmonary-embolism}},
  note={Accessed: 2026-02-22}
}

@misc{drugscom_cortisone,
  title={Cortisone: Uses, Dosage, Side Effects},
  author={{Drugs.com}},
  year={2025},
  howpublished={\url{https://www.drugs.com/cortisone.html}},
  note={Accessed: 2026-02-22}
}

@article{gorelick2002new,
  title={New horizons for stroke prevention: {PROGRESS} and {HOPE}},
  author={Gorelick, Philip B},
  journal={The Lancet Neurology},
  volume={1},
  number={3},
  pages={149--156},
  year={2002},
  doi={10.1016/S1474-4422(02)00071-1}
}

@article{weinblatt1985efficacy,
  title={Efficacy of low-dose methotrexate in rheumatoid arthritis},
  author={Weinblatt, Michael E and Coblyn, Jonathan S and Fox, David A and Fraser, Patricia A and Holdsworth, David E and Glass, Dorothy N and Trentham, David E},
  journal={New England Journal of Medicine},
  volume={312},
  number={13},
  pages={818--822},
  year={1985},
  doi={10.1056/NEJM198503283121303}
}

@article{savarino2018appropriate,
  title={Appropriate use of proton pump inhibitors: when, what, and for how long?},
  author={Savarino, Vincenzo and Marabotto, Elisa and Zentilin, Patrizia and Furnari, Manuele and Bodini, Giorgia and De Maria, Claudia and Pellegatta, Gaia and Coppo, Costanza and Savarino, Edoardo},
  journal={Therapeutic Advances in Gastroenterology},
  volume={11},
  pages={1756284818769936},
  year={2018},
  doi={10.1177/1756284818769936}
}

@article{almusallam2025robust,
  title={A robust deep learning framework for RNA 5-methyluridine modification prediction using integrated features},
  author={Almusallam, Naif and Khan, Salman and Alarfaj, Fawaz Khaled and Ahmad, Nijad},
  journal={BMC Biology},
  volume={23},
  number={1},
  pages={328},
  year={2025},
  publisher={Springer},
  doi={10.1186/s12915-025-02433-2}
}

@article{khan2025integrating,
  title={Integrating AI in security information and event management for real time cyber defense},
  author={Khan, Salman and Dilshad, Naqqash and Ahmad, Nijad and Noor, Sumaiya and AlQahtani, Salman A},
  journal={Scientific Reports},
  volume={15},
  number={1},
  pages={35872},
  year={2025},
  publisher={Nature Publishing Group UK},
  doi={10.1038/s41598-025-19689-x}
}

@article{khan2025deepprobind,
  title={Deep-ProBind: binding protein prediction with transformer-based deep learning model},
  author={Khan, Salman and Noor, Sumaiya and Awan, Hamid Hussain and Iqbal, Shehryar and AlQahtani, Salman A and Dilshad, Naqqash and Ahmad, Nijad},
  journal={BMC Bioinformatics},
  volume={26},
  number={1},
  pages={88},
  year={2025},
  publisher={Springer},
  doi={10.1186/s12859-025-06101-8}
}

\end{document}